# SINGLE PLANE WAVE TAKES A SHORTCUT TO PLANE WAVE COMPOUNDING IN ULTRASOUND IMAGING

*Zhiqiang Li[†], Hengrong Lan[†], Lijie Huang, Qiong He, and Jianwen Luo[*]*

Department of Biomedical Engineering, School of Medicine, Tsinghua University, Beijing, China
[*] E-mail: luo_jianwen@tsinghua.edu.cn

## ABSTRACT

Reconstruction of ultrasound (US) images from radio-frequency data can be conceptualized as a linear inverse problem. Traditional deep learning approaches, which aim to improve the quality of US images by directly learning priors, often encounter challenges in generalization. Recently, diffusion-based generative models have received significant attention within the research community due to their robust performance in image reconstruction tasks. However, a limitation of these models is their inherent low speed in generating image samples from pure Gaussian noise progressively. In this study, we exploit the inherent similarity between the US images reconstructed from a single plane wave (PW) and PW compounding (PWC). We hypothesize that a single PW can take a shortcut to reach the diffusion trajectory of PWC, removing the need to begin with Gaussian noise. By employing an advanced diffusion model, we demonstrate its effectiveness in US image reconstruction, achieving a substantial reduction in sampling steps. *In-vivo* experimental results indicate that our approach can reduce sampling steps by 60%, while preserving comparable performance metrics with the conventional diffusion model.

***Index Terms***— Ultrasound imaging, Diffusion model, Image reconstruction, Plane wave

## 1. INTRODUCTION

Ultrasound (US) imaging is widely used in medical diagnostics, attributed to its noninvasiveness, real-time feature, and cost-effectiveness. Plane wave US (PWUS) imaging has emerged as a research hotspot, especially due to its capabilities to achieve frame rates as high as 10,000 Hz [1]. However, B-mode images reconstructed from PW utilizing the delay-and-sum (DAS) algorithm often exhibit lower image quality due to the absence of transmit focus. To enhance image clarity, PWUS often resorts to plane wave compounding (PWC) with multiple steering angles, albeit at the expense of frame rate reduction. Many methods have been proposed to enhance the image quality of PWUS [2, 3]. In recent years, deep learning (DL) has attracted much attention in US image reconstruction. These models, predominantly trained on paired data [4, 5], have shown to be promise. However, a challenge is their limited ability to adapt to various situations and their tendency to either introduce or amplify specific attributes or features within the learned images [6].

Recent advancements in diffusion generative models have drawn the interest of the research community, primarily due to their state-of-the-art performance, even in situations where paired datasets are not available. Their capability to model intricate, high-dimensional distributions makes diffusion models powerful tools for addressing reconstruction challenges. In image reconstruction problem, the objective often revolves around deducing the original image from given measured data [7-10]. However, a notable limitation of using diffusion models to reconstruct images is their iterative approach to image sampling from Gaussian noise, which leads to substantial computational demands [11].

In this study, our emphasis is on accelerating conditional diffusion models by leveraging the structural similarities in the same imaging region between single PW and PWC images. We find that the forward diffusion trajectories of the single PW and PWC images have a confluence (when noise level $\sigma = \sigma^*$). As they pass through this confluence point (noise level $\sigma \gtrsim \sigma^*$), in the forward diffusion process, the noisy US images reconstructed from PW or PWC follow similar trajectories to Gaussian distribution. This suggests that starting the reverse diffusion process from the original Gaussian distribution (noise level $\sigma = \sigma_{max}$) or from the point after confluence (noise level $\sigma \geq \sigma^*$) would yield similar sampling results. To reduce the number of sampling steps, we introduce perturbations to the single PW image, adding noise to it to reach a noise level $\sigma_k$ ($\sigma^* < \sigma_k < \sigma_{max}$). We then commence the reverse diffusion from $\sigma_k$. This approach utilizes the structural similarities between the single PW image and the PWC image, making it possible to take a shortcut to the PWC image within the reverse diffusion trajectory. Additionally, we incorporate the configurations of diffusion-based generative models (EDM) as delineated in [12], which significantly reduces the total number of steps to 50. Consequently, the total number of reverse diffusion steps required to reconstruct an image through our method is further reduced.

Employing a training dataset comprised of B-modes compounded with 75 PWs, which is denoted as 75 PWs in the rest of paper, we leverage EDM to learn the distribution of the dataset. We demonstrate the effectiveness of our method using *in-vivo* data. In EDM, the low-quality single PW data, the input of the EDM, serves not only as the sampling condition for guiding the reverse diffusion process but also as the basis for initialization sampling as described above. As a result, a high-quality image can be effectively sampled from the learned model in a limited number of steps (i.e., 20 steps). The quantitative evaluations of some results even surpass the image compounded with 75 PWs. Moreover, the performance metrics is close to the that of the EDM result initializing from the original Gaussian distribution.

---

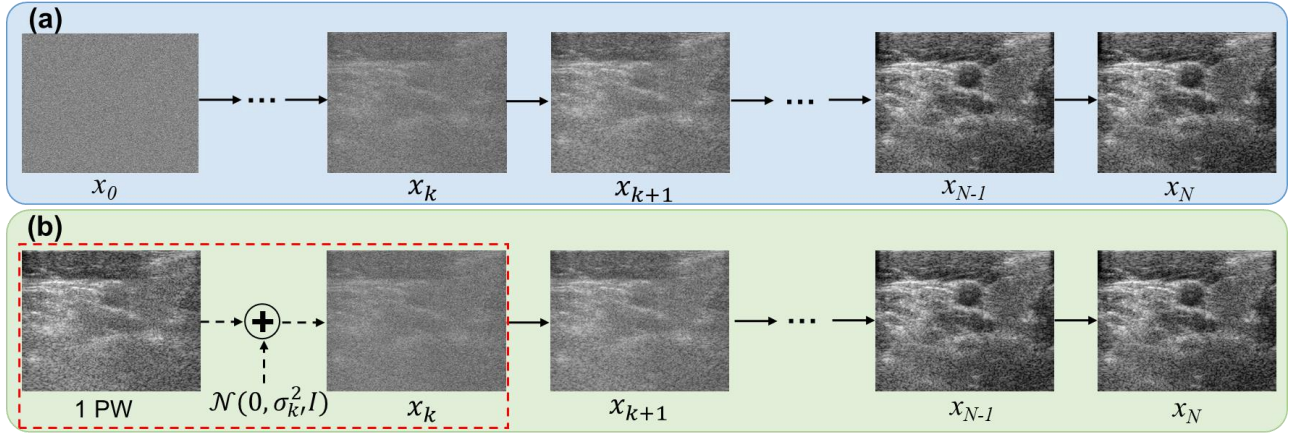

**Fig. 1.** The illustration of sampling methods for PWUS. (a) Conventional diffusion model starts from Gaussian noise $x_0$ (standard deviation $\sigma_0 = \sigma_{max}$) and needs full reverse diffusion. (b) The proposed method utilizes single PW image to accelerate the sampling process, a key component of which is the forward diffusion step of single PW.

## 2. METHODS

### 2.1. Diffusion generative models

The forward diffusion process gradually introduces perturbations to the data $x$ by adding Gaussian noise. At different sample step $t$, the noisy image has varying noise level (standard deviation $\sigma_t$). Throughout this process, a series of mollified distributions, $p(x_t; \sigma_t)$, is obtained from the original data distribution, $p_{data}(x)$. The training process involves a deep neural network (DNN) that predicts the noise based on the input data $x_t$ and the sample step $t$. During the inference procedure, the trained diffusion model can predict the noise at step $t$, which can be used to obtain the noisy images $x_t$ with noise deviation $\sigma_t$ from the noisy image $x_{t+1}$ with noise deviation $\sigma_{t+1}$, iterative ($t \in [0, N]$, $\sigma_0 = \sigma_{max} > \sigma_1 > \cdots > \sigma_N = 0$). In general, the initial noisy image $x_0$ is sampled from $N(0, \sigma_0^2 I)$, as depicted in Fig. 1(a). The desired result of the reverse diffusion is a denoised image $x_t$, which follows the distribution $p(x_t; \sigma_t)$ at each sample step. The endpoint $x_N$ of this process, representing the output of the diffusion model, follows the data distribution $p_{data}(x) = p(x_N; 0)$ as shown in Fig. 1(a).

The stochastic differential equation (SDE) describes the diffusion trajectory of a given data distribution $p_{data}(x)$, as expounded in [11]. Initially, we select a schedule $\sigma(t)$ which defines the desired noise level at a specific time $t$ as described in [12]. In our work, we set $\sigma_{max} = 80$ and $\sigma_{min} = 0.002$ during training process. For the forward process, we can obtain $x_t$:

$$x_t = x_N + \sigma_t N(0, I). \quad (1)$$

This shows that we can calculate arbitrary $x_t$ in straightforwardly thanks to the properties of Gaussian distribution. To represent the solution paths starting from $\hat{x}_0 \sim N(0, \sigma_{max}^2 I)$, we solve the SDE with a numerical method as described in [11]. The endpoint of the trajectory, denoted as $\hat{x}_N$, is considered a sample drawn from the data distribution $p(x_N; 0)$ (i.e., $p_{data}(x)$). Due to iterative evaluations within the DNN, Karras et al. employed the advanced Runge-Kutta method of higher order, resulting in a marked decrease in the number of necessary sampling steps (fewer than 50) [12]. We also employ the method to accelerate the sampling process.

### 2.2. Plane wave ultrasound imaging

For simplicity, the US imaging process is represented as a linear system. In the discretized linear physical model, there are $l$ spatial observation points within the imaging area and $r$ time samples for all $L$ transducer elements. The radio-frequency (RF) signal $y$ received by the transducer can be expressed as $y = Hx + n$, where $y \in \mathbb{R}^{rL}$, $H \in \mathbb{R}^{rL \times l}$, $x \in \mathbb{R}^l$, and $n \in \mathbb{R}^{rL}$. Each colomn of the measurement matrix $H$ represents the received echo signals from the spatial observation point with unifiy scattering cofficient, while $x$ represents the scattering coffcients of the spatial observation point. For simplicity, $n$ is assumed to be white Gaussian noise with a standard deviation $\gamma$, which is reasonable for PW transmission [13].

The beamforming process can be considered as the reverse process of the US imaging process. The beamformed image data $x'$ could be expressed as $x' = B^T y$, where $y \in \mathbb{R}^{rL}$, $B \in \mathbb{R}^{rL \times l}$, $x' \in \mathbb{R}^l$. Each colomn of the beamforming matrix $B$ corresponds to a pixel point, and the elements of each row correspond to the contributions of RF signals $y$. The simplified relationship between the measurement matrix $H$ and the beamforming matrix $B$ is that each colomn of $H$ is a convolution of the transmit waveform with the corresponding colomn of $B$. In this study, we assume that the transmit waveform is a unit pulse, which means that the beamforming matrix $B$ is identical to the measurement matrix $H$ (i.e., $B = H$). Consequently, the beamforming process can be rewritten as $x' = H^T y$. The relationship between the beamforming matrix $B$ and the measurement matrix $H$ is suitable to both single PW imaging and PWC imaging.

Unlike single PW imaging, multi-angle PWC imaging can achieve transmit focus by transmitting multiple PWs with different steering angles and then coherently compounding during the beamforming process. PWC can improve resolution, signal-to-noise ratio (SNR), and overall image quality, but at the expense of frame rate [1]. PWC enhances the main lobe, reduces the sidelobe and incorporates additional information from different angles. Consequently, even when imaging the same region using PWC and PW, there is a structural information difference $\delta$. The relationship between the PWC image $x_c$ and the single PW image $x_s$ can be expressed as:

$$x_c = x_s + \delta - n. \quad (2)$$

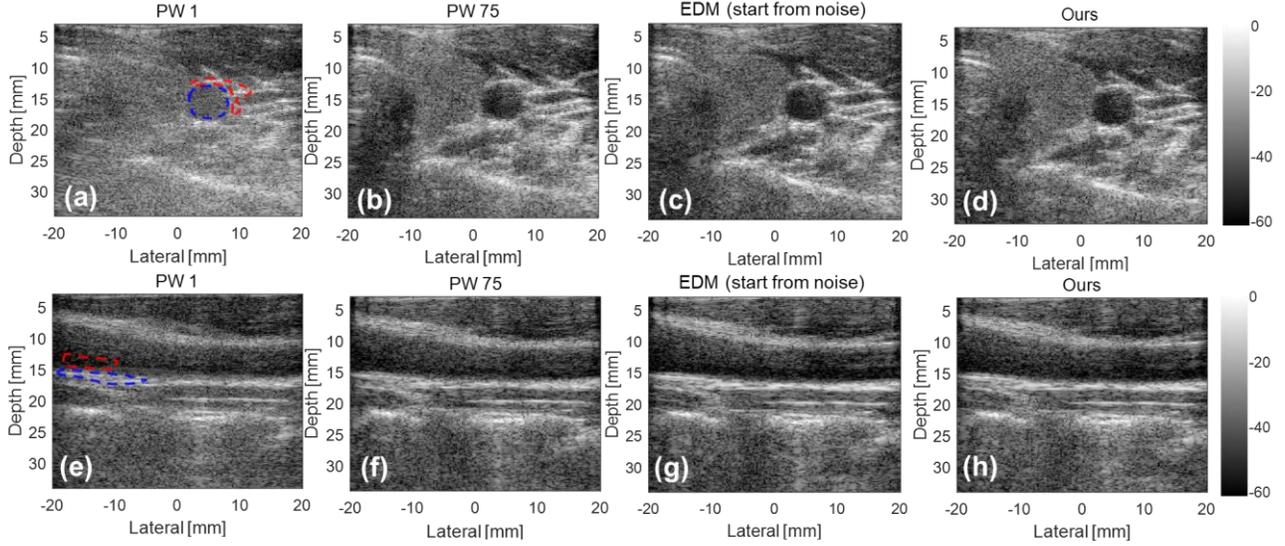

**Fig. 2.** B-mode images of the common carotid artery of a volunteer reconstructed from DAS with [(a) and (e)] single PW and [(b) and (f)] 75 PWs, [(c) and (g)] the sampling results from noise with single PW (50 steps) and [(d) and (h)] the proposed accelerated method with single PW (20 steps), respectively. The [(a)-(d)] first and [(e)-(h)] second rows are from the cross-sectional and longitudinal views of the common carotid artery, respectively.

### 2.3. Accelerated sampling PWC via single PW

In PWUS, we can consider the single PW image as the result of adding noise to the PWC image:

$$x_{k'}^s = x_s + \sigma(k')N(0,I), \quad (3)$$
$$x_k^c = x_c + \sigma(k)N(0,I), \quad (4)$$

This implies that by disregarding the structural difference $\delta$ between the single PW and PWC images, one can derive the relationship by integrating $x_c = x_s - n$ into Eq. (3), yielding:

$$x_{k'}^s = x_c + n + \sigma(k)N(0,I). \quad (5)$$

While the forward diffusion possesses cannot neglect the structural difference $\delta$, the conditional reverse diffusion mitigates this difference at a notably accelerated exponential rate [14]. Notably, it has been validated that the deviation between an initial solution and the true solution during the diffusion trajectory (forward or reverse) diminishes as the superimposed noise intensifies [15]. Consequently, if $\sigma_k^2 = \sigma_{k'}^2 + \gamma^2$, the left-hand sides of the expressions in Eqs. (3) and (4) ($x_{k'}^s$ and $x_k^c$) follow the same distribution with a noise level $\sigma \geq \sigma^*$, which represents the confluence point of the diffusion trajectories of the single PW and PWC images. As a result, there is the noise level ($\sigma \geq \sigma^*$) within the forward diffusion process, which serves as a viable initialization point for the reverse diffusion process of PWC. Therefore, the reverse diffusion process could start with $x_{k'}^s$, as it is on the trajectory of the PWC images.

As Fig. 1(b) shown, our acceleration strategy is to initiate the reverse diffusion from a suitable noise level $\sigma_k$ ($\sigma^* < \sigma_k < \sigma_{max}$), which ensures a substantial reduction in the requisite number of reverse diffusion steps. This can be achieved by adding the Gaussian noise with the standard deviation $\sigma_k$ to the single PW image. It is imperative to note that our method is delineated in **Algorithm 1**. Our reverse diffusion uses Heun's 2$^{nd}$ order method, alternated with a gradient descent operation (step 5 in **Algorithm 1**) to ensure data consistency.

---

**Algorithm 1** Accelerated PWUS reconstruction (EDM)
**Require:** $x_s$, $y$, $H$, $\lambda$, $\sigma_{k'}$, $\sigma_{t \in \{0,...,N\}}$, $k$, $D_\theta$
1:     sample $z \sim N(0,I)$
2:     $x_k \leftarrow x_s + \sigma_k z$    ▷ Forward diffusion
3:     **for** t = k to N-1 **do**    ▷ Reverse diffusion
4:        $\hat{x}_{t+1} \leftarrow$ HeunSampler($D_\theta(x;\sigma)$, $\sigma_{t+1}$, $\sigma_t$)    ▷ Heun 2$^{nd}$ order step
5:        $x_{t+1} = \hat{x}_{t+1} - \lambda \nabla_{\hat{x}_{t+1}}(\|H\hat{x}_{t+1} - y\|_2)$    ▷ Data consistency
6:     **return** $x_N$

---

## 3. EXPERIMENTS

In our study, US images compounded with 75 PWs were employed to train the denoising model, facilitating the learning of data distribution. The EDM framework, realized in Pytorch [16], was trained on a workstation equipped with 128 GB RAM and four NVIDIA TITAN V GPUs.

Our experimental evaluations were executed using *in vivo* datasets. US Imaging was conducted with an L10-5 linear array transducer, connected to a Vantage 256 system (Verasonics Inc., Kirkland, WA, USA). This system captured images composed of 75 PWs, with steering angles spanning uniformly from -16° to 16°. Regarding the *in vivo* dataset, 300 frames of channel RF data were acquired from the right carotid artery of two healthy volunteers, encompassing both longitudinal and cross-sectional views. A test dataset was subsequently created, including 100 frames of the acquired RF data for the reconstruction process. It is noteworthy that all the images reconstructed in this study have dimensions of 1000 × 256 (axial × lateral). For traditional sampling, the maximum time step was set at 50, whereas our innovative approach only required 20 steps.

We have confirmed that using single PW in initialization can accelerate the sampling process of the diffusion model. Because we are unable to quantify the standard deviation γ of *n*, the difference between the single PW and PWC images, we need to determine two parameters, i.e., $\sigma_k$ and $k$. These parameters represent the initial noise level of single PW image and the new maximum noise. We set the initial noise level $\sigma_k = 5$ and compare the performance of different $\sigma_{max}$ with different numbers of steps.

## 4. RESULTS

Figure 2 presents the B-mode images reconstructed with the DAS algorithm with both unaccelerated and accelerated diffusion models. A important observation is that diffusion models provide images with enhanced contrast while maintaining the fidelity of anatomical structures, including the vessel intima. This contrast enhancement is particularly evident when compared to the DAS result with a single PW. When employing DAS with a single PW, the strong noise significantly reduces the contrast of the vessel, making the carotid artery nearly indistinguishable in the cross-sectional view. However, as we increase the number of PWs to 75, there is a noticeable mitigation of this noise. As a result, the carotid artery becomes distinctly visible in both the longitudinal and cross-sectional views. It is of significance to highlight that, even when diffusion models are trained on datasets with 75 PWs, they produce results comparable to thoseof PWC with 75 PWs, despite using only a single PW. Analogous to results generated by diffusion model with initialization from Gaussian noise, our approach also achieves comparable performance. It is important to mention that while the EDM framework reduces the iteration count to 50, our accelerated approach requires only 20 iterations.

**Table 1.** Quantitative results (CNR in dB and gCNR) of the *in vivo* results for different methods.

|  |  | Single PW | PWC 75 PWs | EDM 1 PW | Ours 1 PW |
|---|---|---|---|---|---|
| Cross-sectional | CNR | 1.0363 | 1.6339 | **2.1500** | 1.9962 |
|  | gCNR | 0.6574 | 0.8408 | 0.8721 | **0.8955** |
| Longitudinal | CNR | 5.1481 | **6.1987** | 6.0090 | 6.1710 |
|  | gCNR | 0.9683 | 0.9897 | 0.9935 | **0.9938** |

Regions of interest (ROIs) were selected within the vessel wall and the carotid artery lumen (highlighted by red dashed lines) in both the longitudinal and cross-sectional views. This allows to calculate both the contrast-to-noise ratio (CNR) and the generalized contrast-to-noise ratio (gCNR) [17], as presented in Table 1. The diffusion models, when applied to a single PW, yield CNR and gCNR values that are similar to those achieved by PWC with 75 PWs.

The aforementioned results are based on specific parameters ($\sigma_k = 5$, $k = 30$, $\sigma_{max} = 60$). To further explore the impact of these parameters on the quantitative results, we examined the effects of the iteration count $k$ and $\sigma_{max}$ value on the mean gCNR of the two ROIs while keeping $\sigma_k$ constant at 5, as shown in Fig. 3. Interestingly, regardless of the $\sigma_{max}$ value, there is a threshold for the iteration count beyond which the results surpass those of PWC with 75 PWs. Figure 3 further suggests that the iteration count can be reduced to 10, with only a marginal performance decrease when compared to the results with 20 iterations. It is vital to note that selecting $\sigma_{max}$ and $\sigma_k$ is not straightforward due to the absence of a definitive relationship between single PW and PWC images.

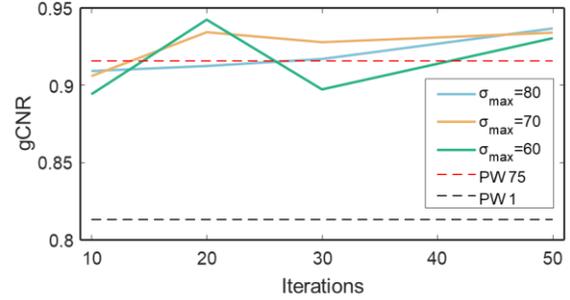

**Fig. 3.** Comparison of different $\sigma_{max}$ using $\sigma_k$ =5. For each curve, we also compare the gCNR of different iteration counts s. Black and red dashed line are single PW image and PWC with 75 PWs, respectively.

## 5. CONCLUSION

In this work, we introduce an accelerated EDM-based reconstruction technique to enhance the image quality of PWUS. In addressing reconstruction via conditional reverse diffusion, we diverge from the conventional approach that starts with noise. Instead, we propose to initiate the process from a forward-diffused single PW image. Our experimental evaluations demonstrate the effectiveness of this 'shortcut' trajectory from the single PW to PWC images. This method not only accelerates the sampling process but also improves the stability and performance of PWUS. As a potential future direction, we expect the development of an adaptive mechanism capable of automatically selecting the optimal values of $\sigma_{max}$ and $\sigma_k$.

## 6. COMPLIANCE WITH ETHICAL STANDARDS

Approval of all ethical and experimental procedures and protocols was granted by the Medical Ethics Committee of Tsinghua University.

## 7. ACKNOWLEDGMENTS

The authors would like to express their thanks to Shangqing Tong (at ShanghaiTech University) for his invaluable support and contribution towards the methodology and discussion of this paper. This work was supported in part by the National Natural Science Foundation of China (62171442 and 62027901) and Beijing Natural Science Foundation (M22018).